\documentclass[sigconf]{acmart}

\usepackage{balance} 

\AtBeginDocument{%
  \providecommand\BibTeX{{%
    \normalfont B\kern-0.5em{\scshape i\kern-0.25em b}\kern-0.8em\TeX}}}

\copyrightyear{2023}
\acmYear{2023}
\setcopyright{acmlicensed}\acmConference[K-CAP '23]{Knowledge Capture Conference 2023}{December 5--7, 2023}{Pensacola, FL, USA}
\acmBooktitle{Knowledge Capture Conference 2023 (K-CAP '23), December 5--7, 2023, Pensacola, FL, USA}
\acmPrice{15.00}
\acmDOI{10.1145/3587259.3627551}
\acmISBN{979-8-4007-0141-2/23/12}

\begin{document}

\title[Finding Neural Concept Representations]{Finding Concept Representations\\in Neural Networks with Self-Organizing Maps}

\author{Mathieu d'Aquin}
\email{mathieu.daquin@loria.fr}
\orcid{0000-0001-7276-4702}
\affiliation{%
  \institution{LORIA, Université de Lorraine/CNRS/INRIA}
  \city{Nancy}
  \country{France}
}

\renewcommand{\shortauthors}{M. d'Aquin}

\begin{abstract}
In sufficiently complex tasks, it is expected that as a side effect of learning to solve a problem, a neural network will learn relevant abstractions of the representation of that problem. This has been confirmed in particular in machine vision where a number of works showed that correlations could be found between the activations of specific units (neurons) in a neural network and the visual concepts (textures, colors, objects) present in the image. Here, we explore the use of self-organizing maps as a way to both visually and computationally inspect how activation vectors of whole layers of neural networks correspond to neural representations of abstract concepts such as `female person' or `realist painter'. We experiment with multiple measures applied to those maps to assess the level of representation of a concept in a network's layer. We show that, among the measures tested, the relative entropy of the activation map for a concept compared to the map for the whole data is a suitable candidate and can be used as part of a methodology to identify and locate the neural representation of a concept, visualize it, and understand its importance in solving the prediction task at hand.
\end{abstract}

\ccsdesc[500]{Computing methodologies~Artificial intelligence}
\ccsdesc[500]{Computing methodologies~Neural networks}
\ccsdesc[500]{Computing methodologies~Knowledge representation and reasoning}

\keywords{Neural networks, conceptual representation, neuro-symbolic AI}



\maketitle

\section{Introduction}

It has now been clearly established that neural networks represent an effective, popular, and highly applicable approach to data-centric artificial intelligence, but that one of their key disadvantages is their interpretability~\cite{von2021transparency}. They lack transparency in the sense that, even if one can inspect their inner working, a meaningful understanding of the relation between the (sometimes extremely large) vectors of weights and activations in the network and the conclusion being made is rarely achievable. However, some form of implicit representation of abstract notions, conceptual knowledge, is expected to exist within those vectors to support the task at hand~\cite{net2vec}. For example, in a previous work~\cite{cikm2020}, we could see that there was a strong relationship between the results of neural networks and the presence concepts extracted from a knowledge graph in the input data, showing, for example, that a network appeared to rely on knowledge of the country of origin or artistic movement of a painter to predict whether their work was exposed in major museums.

In this paper, we aim to explore a methodology to identify and locate the representation of conceptual knowledge directly within a neural network's layers, to find for example whether the concepts of `Italian painters' or 'surealist painters' are actually present in the activation vectors of the network. In other words, we aim to propose a way to inspect those vectors so that they can be visually and computationally assessed and compared when presented with input data exemplifying different concepts. This can help to understand the importance, direct or indirect, of certain concepts in the decision and where the representation of those concepts might be located in the network. This is helpful in order to not only make the neural network more interpretable and communicable, but also to potentially identify biases in the data and in the way the network exploits the data. Indeed, in such cases, we can see biases as the use of information that should not be relevant to make a decision. By identifying a supposedly irrelevant concept that is implicitly represented in a neural network and how close the representation of such a concept is to the decision taken (i.e., to the output layer), we can therefore get an idea of the level at which that particular form of bias is present within the network. 



To achieve that, we use self-organizing maps (SOMs~\cite{refsoms}). SOMs are neural network architectures using competitive learning to create a low-dimensional map (generally a 2D grid) of highly dimensional data. It is often used for tasks such as clustering, with the units on the grid representing clusters organized based on their similarity. Here, we use SOMs to provide abstractions of the activation vectors of a neural network from which we expect it to be easier to spot patterns and regularities. We test a number of metrics on the constructed activation maps to identify which could be used to provide an assessment of the level of representation of a concept within a neural network layer and show how the selected measure behaves on two example neural network models: one (smaller scale) used for text classification and one (larger) used in a regression task from images. We show that our method can successfully identify the relative importance of the representation of concepts such as the ones associated with DBpedia categories, the gender of a person or their ethnicity, in the different layers of the neural networks.  

\section{Related Work}

This work is strongly related to the explanation and interpretation of neural networks, for which there are many techniques~\cite{xaisurvey}. Some approaches use knowledge graphs (see~ \citet{ilariasbook} for an overview), but most explanation techniques focus on identifying the features (i.e., the part of the input) that have had the greatest impact on the output. Although this has shown to be helpful in identifying biases in the data involved in training~\cite{cleverhans}, approaches based on this idea cannot, by nature, help to identify concepts of higher level, such as the gender of the represented person. We show below that, through our method, it is possible to visualize and assess the level of representation of characteristics such as the gender, artistic movement or ethnicity of a person.

In addition to focusing on input features, many of the shortcomings of explanation and interpretation approaches come from the fact that they are external to the neural network and therefore do not really help to understand the decision process involved, only its behavior. In other words, they process the input and output of the network without looking at its inner working. For that reason, researchers have started looking at how analyzing neuron activations in a network can help to extract information about the way it reaches a conclusion, for example by processing networks of co-activations of neurons~\cite{tiddimileo}. 

Closer to the work presented here, researchers have tried to identify specific units (neurons) that appear to activate when specific `concepts' are present in the data~\cite{cnnvis}. While such works could find correlations between concepts and neural activations, it is expected that the neural representation of abstract concepts would receive contributions from multiple units (or filters in vision)~\cite{net2vec}. In addition, most of the works in this area are focused on visual concepts: textures, patterns, or a particular type of objects present in an image, and are designed to work only on convolutional neural networks (CNNs). We aim here to be able to identify the presence of representations of more abstract concepts from all kinds of data, in all kinds of network (e.g. the artistic movement in a recurrent neural network trained on textual data about painters). 

In~\citet{TCAV}, although they still focus on visual concepts (e.g. `stripes on a zebra'), the authors propose a more general approach (Testing Concept Activation Vectors, TCAV) to identifying whether `concepts' appear in neural network. While there are many similarities between our work and TCAV, we explore here an alternative technique more suited to the visual exploration of the representation of concepts in the network. Finally, another recent work~\cite{HHAI2023} also tested the use of SOMs (among other approaches) to visually represent activations in a network. It showed that such a representation could be suitable to inspect neural networks (already trained or during training), for the purpose mostly of visual inspection and model debugging, and without considering the way externally defined concepts could appear in those activations. 

\section{Building Self-Organizing Maps from Neural Network Activations}

In this section, we describe the proposed method for identifying and localizing representations of concepts in layers of neural networks using self-organizing maps. We start by informally introducing self-organizing maps and providing an overview of the process to build, compare, and assess SOMs from neural networks' activation vectors (activation maps). We then introduce an example, on which we will rely in the next section to test different measures of the level of representation of concepts in activation maps. 

\subsection{Self-Organizing Maps}

SOMs are a family of neural network architectures mostly used for unsupervised learning, that is, to identify patterns and structures in potentially high-dimensional data. A typical SOM is a rectangular grid (although other shapes are possible) of units, all connected to the input. During training, the activation of each unit is a function of the similarity of the units' weight vector to the input vector. The weights are then updated following a competitive learning method in which the most similar unit (the winning neuron) is first selected, and its weights (as well as, to a lesser extent, the weights of the units in its neighborhood) are updated to increase the similarity with the input example. By iterating through the training data in this way, a self-organizing map will tend to converge towards one where hot-spots of similar data points will be grouped in particular areas of the map, and where distances on the map reflect distances in the higher-dimensional space of the input data. In this sense, SOMs can be seen as both a clustering/discretization method and as a dimension reduction method. 

In the implementation of the process described in this paper, we use the library \emph{minisom}\footnote{\url{https://pypi.org/project/MiniSom/}} to build SOMs from activation vectors. While \emph{minisom} might not be the most efficient alternative, it is well established, robust, and highly parameterizable and, therefore, suitable for the needs of the research prototype developed as part of this research. A more efficient implementation that can, in particular, rely on GPUs will be integrated in future evolutions.

\subsection{Overview of the process}
\label{sec:overview}

\begin{figure*}
    \centering
    \includegraphics[width=\textwidth]{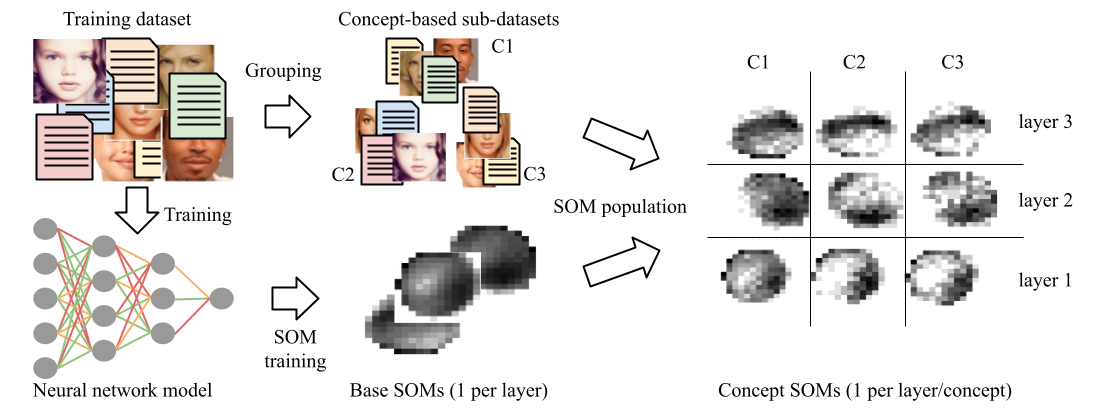}
    \caption{Overview of the process: Starting from a neural network trained on a given dataset, we first build SOMs based on the activation vectors of each layer of the network on that dataset. We then populate those SOMs using the activation vectors in the neural network from subsets of the dataset that correspond to specific concepts. The SOM activation matrices obtained can then be assessed and compared.}
    \label{fig:overview}
\end{figure*}

An overview of the overall process for analyzing the presence of concept representations in the activations of a neural network models using SOMs is presented in Figure~\ref{fig:overview}. The code used in this paper to carry out this process in two case studies is available at \url{https://github.com/mdaquin/actsom}.

This process assumes the availability of a trained neural network model, of the dataset on which it was trained, and of concept labels for the elements of this dataset. In principle, there are no restrictions on the architecture of the network used, as long as activation vectors can be extracted from it. In the use cases presented below, we apply it to networks containing fully connected, convolutional, and recurrent (LSTM) layers. The process also assumes that the dataset is labeled with relevant concepts, i.e. that it is possible to create sub-datasets containing examples (data points) that are representative of the concepts we aim to assess. 

The first step in the process is to create SOMs based on the activation vectors of each layer of the network when applied to the entire dataset. This corresponds to using as input to the SOM training process for a given layer the set of activation vectors for that layer that are extracted when running a forward pass on each example in the dataset. Each of those SOMs is trained (through the \emph{minisom} library) using the typical competitive learning approach. Here, since our objective is not to fully analyze the effects of those parameters, we fixed the values of hyperparameters based on preliminary tests, using a cosine distance as the metric for activation, the `Mexican hat' neighborhood function, a neighborhood radius (sigma) of 8, and SOM grids of $15\times15$ units. Also, for layers of more than one dimension, we specify the aggregation function to use in order to transform their activation matrices or tensors into activation vectors. In the use cases below, we aggregated using the mean of activation values for recurrent and embedding layers, and by flattening the activation tensor for convolutional layers. 

We call the resulting SOMs the `base SOMs' as they correspond to the patterns of activation in each layer when all concepts are present, and therefore will form a base for comparison. We represent those SOMs by a matrix of the activation frequency of each unit in the SOM on the entire dataset (i.e., for a given unit, the frequency at which it was the winning unit for the examples in the dataset). This matrix is represented visually as a heatmap, as visible in Figure~\ref{fig:overview}, where darker tones represent higher frequencies. 

Once the base SOMs are built, we build similar matrices (and heatmaps) for each of the subsets of the dataset corresponding to a concept. In practice, this simply means that, for a given layer and a given concept, we activate the base SOM corresponding to the layer for all examples of the concept (i.e., we `populate the map') and build the corresponding frequency matrix. The idea is that those `concept maps' can then be analyzed and compared to the corresponding base SOM (as well as to each other) to find whether specific patterns are present in the activations for a given concept. Our goal here is, in particular, to find what metrics could be used to assess a given concept map to understand how well the corresponding concept is represented in the corresponding layer.

\subsection{Illustration on an Example}

As a first use case and the basis for our tests, we use an example similar to that used in our previous work~\cite{cikm2020}. All the code to collect the dataset, the concept annotations for the dataset and to train the model (as input to the process presented above) is available in the provided online repository. 

In this use case, the objective of the model is to predict, given the biography of a painter, whether this painter has at least one painting displayed in at least one major museum. Data are obtained by querying DBpedia\footnote{\url{https://dbpedia.org}} for the abstracts of entities of type \emph{painter}, together with the number of museums that refer to objects related to the given painter. We train a simple recurrent network made of an embedding layer, an LSTM layer, and a fully connected layer with relu as activation function in a binary classification task (i.e., with two classes: the painters with at least one painting in a major museum and the ones without). After a few trial-and-error-based tests to identify good values for hyperparameters and balancing the dataset through undersampling, we obtained a network that achieved $72.6\%$ accuracy on the test set (20\% of the balanced dataset).\footnote{The original dataset contained 23.8K painters, of which only 2,416 had paintings in major museums. For undersampling, we therefore randomly selected 2,416 painters who did not.} While there is no doubt that better results could be obtained, our objective here is to understand what subconcepts of `painters' might have been identified and represented in the network to support its predictions, which is useful even if those predictions are not very accurate (possibly even more).

Concerning the concepts to be considered, we also retrieve at the time of querying DBpedia, for each painter, their nationality, the artistic movement(s) to which they belong, and their categories (according to the DBpedia category taxonomy). This allowed us to group painters in our dataset according to concepts such as `Italian Painters', `Realist Painters', or `American Women Painters', in addition to the target classes, i.e. painters with paintings in major museums and painters without. 

\begin{figure*}[t]
    \centering
    \includegraphics[width=\textwidth]{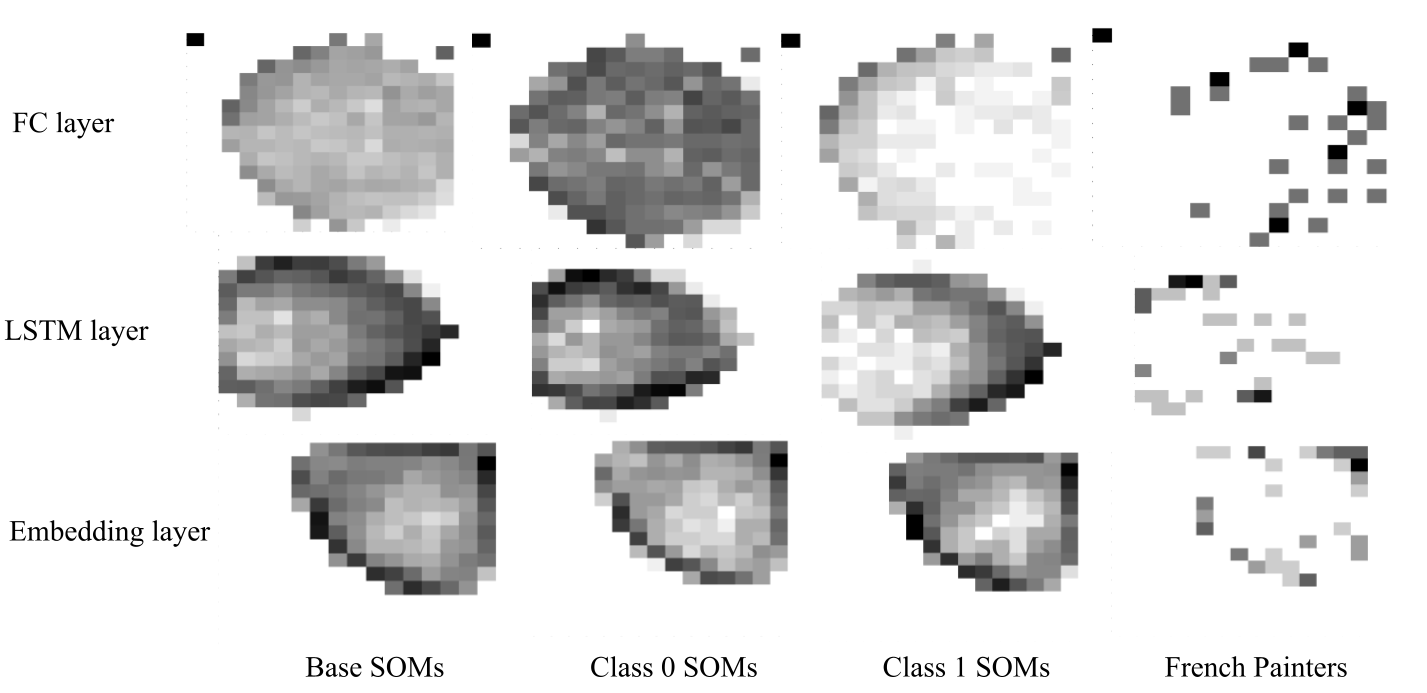}
    \caption{Base SOMs, concept SOMs for the target classes and concept SOMs for painters of French nationality for each layer of the model to classify painters.}
    \label{fig:painters}
\end{figure*}

Figure~\ref{fig:painters} shows the base SOMs produced for each of the three layers of the model, as well as the SOMs corresponding to the target class~0 (painters with no painting in major museums), to the target class~1 (painters with paintings in major museums) and to the concept of French painters (painters of nationality `French'). When comparing those SOMs, clear differences can be seen. It is in particular visible that activations for the two target classes are substantially different in the fully connected layer (FC, the last one before the output layer in the network), but not so much in the previous layers (which is something we will use in the next section). Focusing on the concept of French painters, we can see that the heatmaps for the concept SOMs are more scattered than the base ones, which is unsurprising considering that French painters correspond to a small subset (151 painters) of the overall dataset (23.8K painters). However, we can also see that the distribution of frequencies in the units of the SOMs might show a pattern indicating that the concept of French painters has a stronger representation in the FC layer (many darker units in the lower right corner) than it does in the embedding layer (in which darker units tend to be situated where they are also darker in the base layer). Therefore, the key question here is: How to assess the level of representation of a concept in each layer more formally, beyond the visual inspection of the heatmaps? 

\section{Assessing the Level of Representation of a Concept}

In this section, we show the results of testing different measures in the use case presented above to better understand which metrics could be used to assess how well a concept is represented in a given layer of a neural network. We discuss the measures selected based on how they could intuitively help identify well-represented concepts and test those measures applied to the SOMs as represented by their activation frequency matrices (as in the heatmaps in Figure~\ref{fig:painters}).\footnote{We also tested those measures on matrices corresponding to the average activation values of the SOM, i.e., the average distance of each tested example to the weight vector of the units. However, the results were either similar to the ones when using the frequency maps, or provided values ranging in very small intervals, rendering them difficult to interpret. Therefore, we omit the results for those here.} 

Understanding whether a measure is a good indicator of the level of representation of a concept in the activation vectors of a neural network is not trivial. To achieve this, in our tests below, we rely on a reasonable hypothesis: \emph{If we consider that the target classes of the model correspond to concepts, such concepts should be better represented in the layers closer to the output of the network than they are in those closer to the input layer}. This hypothesis is based on the idea that the different layers of the network have a role to progressively abstract the examples provided so that they can be classified into the target classes in the end (in our example, painters with their work in major museums and painters without their work in major museums). This hypothesis can be visually verified in Figure~\ref{fig:painters} where the SOMs for the target classes are progressively more distinct from the corresponding base SOMs moving from the embedding layer to the FC layer, as previously mentioned. 

In addition to testing, for each measure, whether they validate this hypothesis in the presented use case, we consider two requirements for those measures: 1- that they should have values in intervals that facilitate comparisons, and 2- that the evolution of their values through the different layers should be consistent across the two target classes. 

\subsection{Tested Measures}

We consider two categories of measures here, depending on whether they apply directly to a SOM (or, more precisely, to the activation frequency matrix for a concept in a SOM) or whether they are measures comparing a concept SOM to a base SOM for the same layer. 
  
\subsubsection{Base measures}$ $\\[-3mm]

\textbf{Entropy:} SOMs tend to group together in a given area of the map examples (in our cases, activation vectors) that are similar. Therefore, the frequency matrix for a well-represented concept would be expected to have high values concentrated within a small area of the map. Shannon's entropy is a measure of the level of uncertainty associated with the outcomes of a random variable. Here, considering the unit active in the SOM as the random variable, it would thus measure how close to a random distribution the frequency values in the concept SOM are. In order to obtain high values for high levels of representation, we use the inverse of entropy, i.e. if $s$ is the frequency matrix for a concept SOM represented as a vector of probabilities for each unit to be activated, and $E(x)$ is Shannon's entropy applied to a probability distribution $x$, then the measure used is $1/E(s)$.\\

\textbf{Max FM:} Another intuition on which we could rely is that if a concept is well represented by a layer, then many of the instances of the concept will activate the same unit of the SOM, which would also not activate for examples outside of the concept. We can measure how representative each unit of the SOM is of a concept by using the typical F-Measure from information retrieval. The measure used here is therefore, for a given SOM and a given concept, the maximum F-Measure, i.e. the F-Measure of the SOM unit that is most representative of the concept. 

\subsubsection{Measures comparing concept SOMs and base SOMs}$ $\\[-3mm]

The frequency matrices for the base SOMs of each layer of the model can be considered as the average patterns throughout the dataset. It is therefore meaningful to consider measures that assess how specific concept SOMs diverge from the corresponding base SOMs to assess how activation vectors in the neural network are specific to the concept. \\

\textbf{Distance:} This measure corresponds, essentially, to the cosine distance between a concept SOM (the frequency matrix) and the corresponding base SOM. \\

\textbf{Relative Entropy:} The relative entropy (or Kullback–Leibler divergence) measures how a given probability distribution $P$ diverges from an expected or reference probability distribution $Q$. Here, we use the frequency matrix of the concept SOM as the tested probability distribution $P$, and the frequency matrix of the corresponding base SOM as the reference distribution $Q$. In other words, this measures how unexpected the patterns of activation in the concept SOM are if we expect the ones in the base SOM.

\subsection{Results}

\begin{figure}
    \centering
    \includegraphics[width=1.1\linewidth]{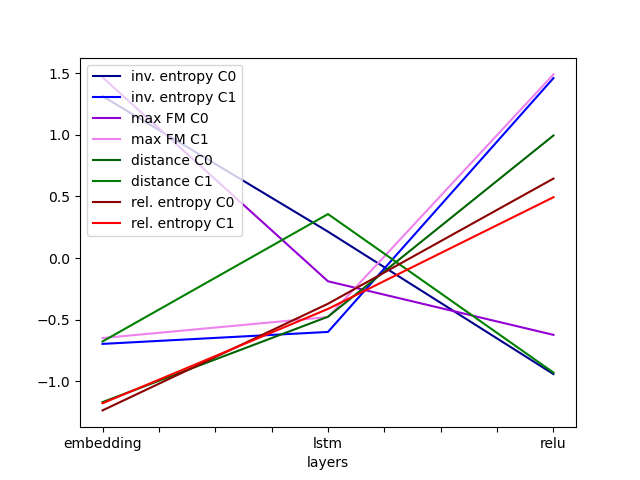}
    \caption{Evolution of measures of the level of representation of concepts applied to target classes across layers of the network. `relu' corresponds to the fully connected layer.}
    \label{fig:measures}
\end{figure}

Figure~\ref{fig:measures} shows the evolution of the values of each measure as applied to the concept SOMs corresponding to the target classes of the model (museum or not museum). The values of the measures were standardized so as to bring them to the same scale (i.e. the z-score for each value is shown). 

As can be seen, in the majority of cases, the expected behavior is shown: The value of the measure increases as we get closer to the output layer of the network. However, for all other measures than relative entropy, the values evolve very differently for the two concepts under consideration (C0 and C1), therefore failing to meet our consistency requirement. Relative entropy, on the other hand, is consistent and clearly increases from the embedding layer to the FC layer (named `relu' in the figure). In addition, it ranges in this case between 0.039 and 0.52, showing that it is suitable as a measure for comparison. 

It can also be noticed that there seems to be a correlation between the entropy and the max FM measure, although it is unlikely to be significant. 

\section{Use Case: Analyzing the Most Represented Artistic Movements and Categories of Painters}

To check whether the analysis of the activation SOMs as computed following the process presented above could be useful, we consider two use cases, the first of which relies on the model and dataset previously introduced to illustrate the process and test measures. As mentioned earlier, we collected additional information about the painters from DBpedia, which was not used when training the model, including the artistic movement to which they belong and their category in the DBpedia category taxonomy. 


\begin{figure}[h!]
    \centering
    \includegraphics[width=1.1\linewidth]{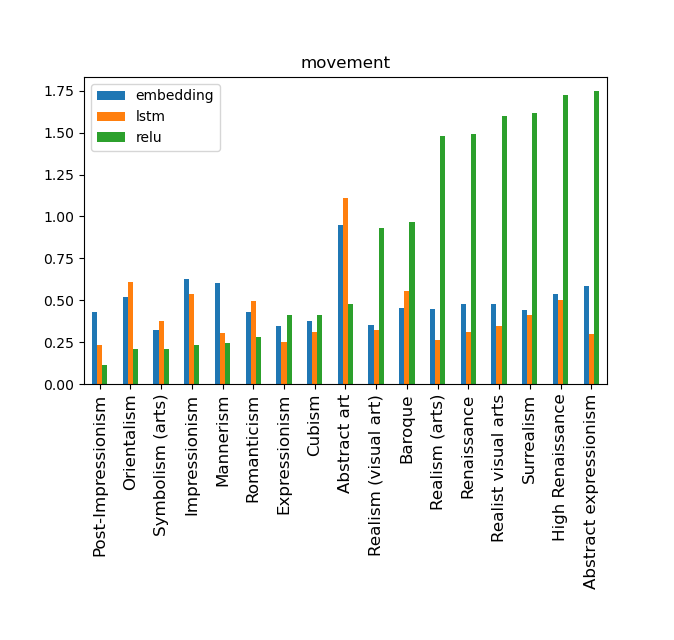}
    \caption{Relative entropy of concepts related to the artistic movement of painters.}
    \label{fig:movement}
\end{figure}

As a first test, we calculated the relative entropy of the SOMs for each layer for the artistic movements related in the dataset to more than 20 painters. The results are presented in Figure~\ref{fig:movement} in order of relative entropy for the fully connected (relu) layer. As can be seen, some concepts such as the one of `painters of the abstract impressionism movement' appear to be significantly more represented in the relu layer than others, while there representation appears to be average in the other layers. This could possibly indicate a connection between those concepts and the decision. Indeed, as we have seen, the relu layer being the last one before the output layer, it is the one where the target classes should be highly represented. Therefore, the fact that a particular concept is highly represented in that layer might mean that this concept overlaps strongly, at least at a conceptual level, with the target classes. On the other hand, we can also here identify concepts, such as the one of `painters in the abstract art movement' that are strongly represented in the layers closer to the input, which would indicate that the language and vocabulary used to describe those painters might be significantly different from those for the others, without this necessarily having a strong impact on the prediction (since the same concept has a relatively low representation in the relu layer). This demonstrates how looking at the level of representation of concepts in the layers of the neural network can help to better understand how the model uses (and abstracts from) the dataset to make its predictions. 

\begin{figure*}
    \centering
    \includegraphics[width=\textwidth]{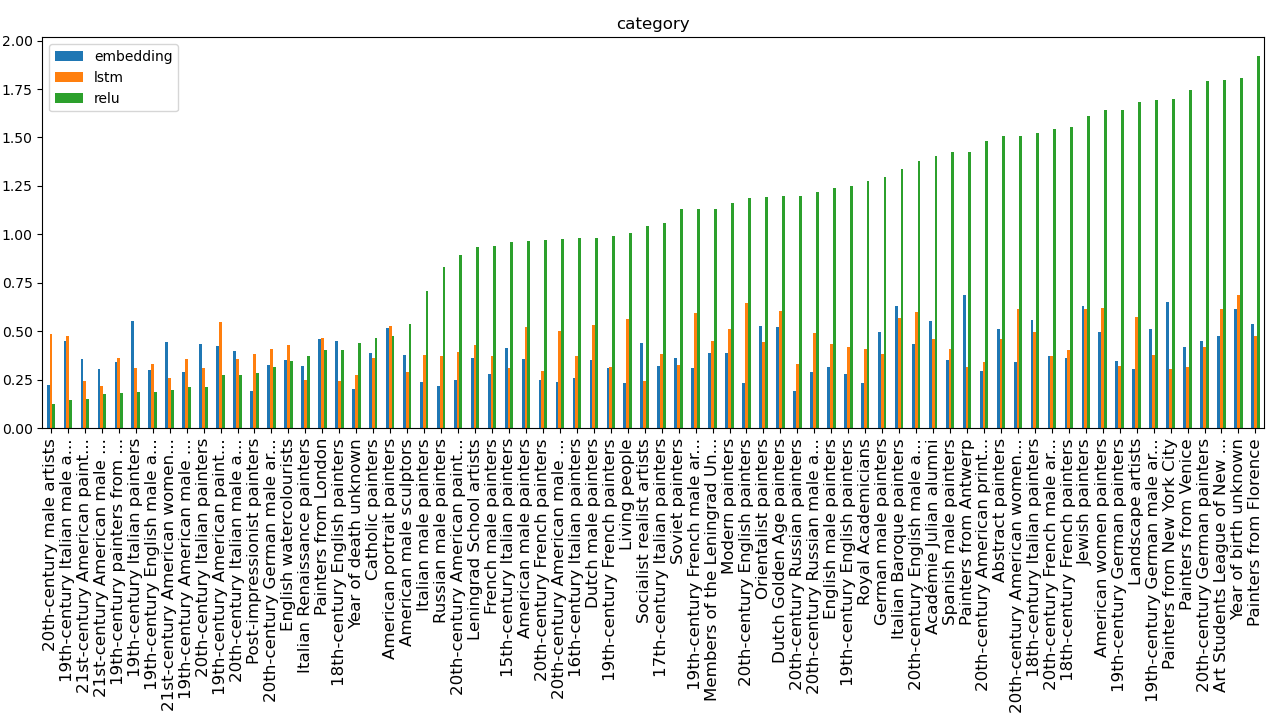}
    \caption{Relative entropy of concepts related to the DBpedia categories of painters.}
    \label{fig:categories}
\end{figure*}

Applying the same process on the categories of painters (taking those related to at least 30 painters) instead of the artistic movements leads to much more varied concepts, as seen in Figure~\ref{fig:categories}. Here too we can observe that some concepts, such as `painters from Florence' are better represented in the relu layer, potentially showing either a rule in the domain, or a bias if that rule turned out not to reflect reality (e.g. that painters from Florence tend not to have their paintings in museums). Similarly as with the artistic movement, other concepts are more presented in the initial layers, which are more focused on extracting features from the data (in this case, language features). The concepts of `painters from Antswerp' and `painters from New York City' for example are relatively strongly represented in the embdedding layer, indicating that the biographies of painters belonging to those concepts are likely to use a significantly different vocabulary from others. Interestingly, many of the highly represented concepts here relate to locations, as well as other personal attributes of the painters (e.g. `American Women Painters' or `Jewish Painters') rather than to the characteristics of their works (e.g., `Landscape Artists'). Considering that this is a small model trained on data of rather poor quality and not reaching a particularly high accuracy, it seems consistent that predictions from that model would rely on the representation of crude concepts providing indirect and potentially biased predictors for the target classes (confirming some of the results obtained previously on a similar dataset in~\cite{cikm2020}).

\section{Larger-Scale Use Case: The representation of gender and ethnicity in predicting a person's age from a photo}

As discussed above, our first use case is an interesting demonstration of the potential of the method presented here, as it allows us to get a better understanding of the way the prediction can be based on exploiting shortcuts and biases in the data. To test this method on a model with a different architecture and of a larger scale, we apply it on one trained to predict the age of a person from a photo. For this, we reused the ResNet18 vision model~\cite{resnet18} pretained on the ImageNet dataset, with an additional fully connected layer, and trained it on the UTKface dataset\footnote{\url{https://susanqq.github.io/UTKFace/}}. We obtained a model that achieved a $R^2$ score of $80\%$, corresponding to a mean absolute error of $6.44$ years. An advantage of the UTKface dataset is that each image is also annotated with the gender and ethnicity of the person. While we do not use those annotations in training it, they allow us to test how the corresponding concepts are represented in the model. All the code to train the model, extract the SOMs and test them is available in the provided online repository. 

\begin{figure}
    \centering
    \includegraphics[width=1.1\linewidth]{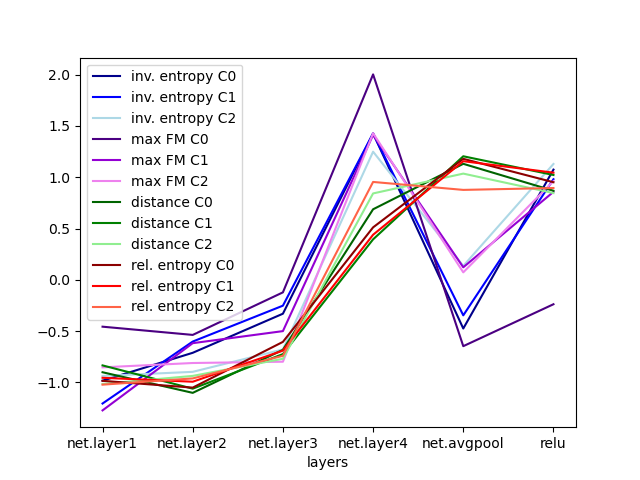}
    \caption{Evolution of measures of levels of representation of concepts applied to age groups across layers of the network.}
    \label{fig:measures2}
\end{figure}

To check the validity of our choice of the relative entropy as a measure of the level of representation of a concept, we discretized the target values into three groups (young, medium, and old) using the k-means method and tested the evolution of our four measures across the layers of the model (net.layer1, net.layer2, net.layer3, net.layer4 and net.avgpool being the successive layers of ResNet18 and relu being the added fully connected layer). As can be seen in Figure~\ref{fig:measures2}, relative entropy continues to broadly meet the requirements expressed earlier. Interestingly, in this case, all the measures are significantly more consistent, even though they do not all evolve in the way expected (except for distance) as we move up the different layers.


\begin{figure*}
    \hspace*{-0.1cm}\includegraphics[width=0.5\textwidth]{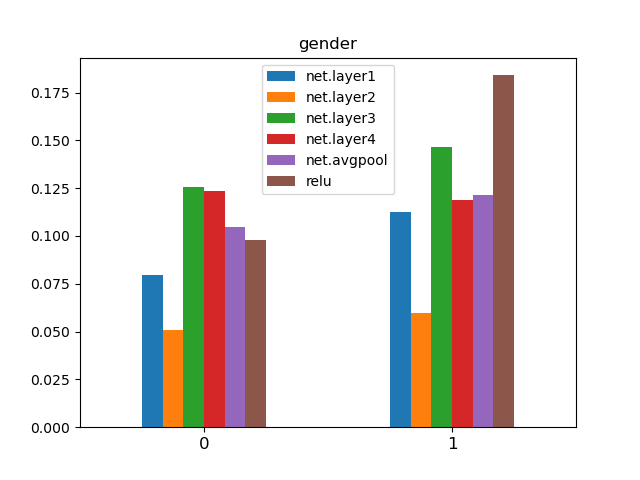}
    \includegraphics[width=0.5\textwidth]{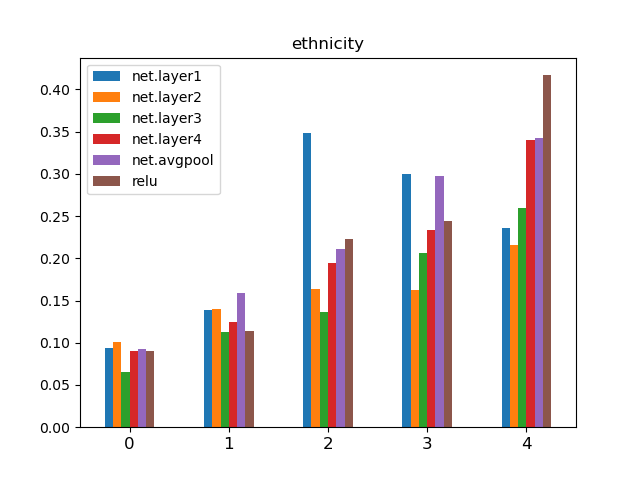}
    \caption{Relative entropy of concepts related to gender and ethnicity in the age prediction use case.}
    \label{fig:agenet_rel}
\end{figure*}

Having verified that relative entropy remains a valid measure here, our goal is to use the proposed method to check how much the concepts corresponding to the different genders are represented and where in the network those representations are stronger. Figure~\ref{fig:agenet_rel} therefore shows the values of relative entropy for those concepts in the different layers of the network. As can be seen, the representation of the different genders and ethnicity appear to evolve consistently across the layers with some exceptions. In particular, both genders are clearly less represented in layer~2 of the ResNet18 component of the network, but the gender labeled 1 has a slightly higher representation generally in all layers, and is significantly more represented in the last (relu) layer. Similarly, the concepts corresponding to the different ethnicities have different levels of representation but follow roughly a similar trend. Ethnicity~2 is however more clearly represented in the relu layer (and therefore could represent an important concept for the prediction), while ethnicities 1.0 and 4.0 are more represented in the lower level layer net.layer1, indicating that they might be visually more distinguishable than others in the data. It is useful here also to notice that, on average, the concepts corresponding to ethnicities have a higher relative entropy across layers than the ones corresponding to gender, which seems to indicate that ethnicity is both visually and conceptually more important to the model when predicting age than gender. We, of course, cannot rule out that this might be due to biases in the datasets (e.g., different age distributions for the different ethnicities). 

\section{Conclusion}

In this paper, we explored how building and comparing SOMs on the activation vectors of layers of neural networks can help analyze the relative level of representation of concepts such as those associated with artistic movements, location, gender or ethnicity in a neural network model. We found in particular that the relative entropy of the probability distribution of unit activation in a SOM for a given concept, compared to that distribution for the whole dataset, provides a useful measure to check how much a concept is represented within a layer. Through two use cases, we showed that this method and measure can help better understand the basis for predictions in a network and potential biases on which it might rely. In other words, in addition to providing a method to assess the presence of a neural representation of concepts at different levels of a network, our test validated the idea of using SOMs to abstract patterns of activation into structures that are computationally and visually processable. Beyond what is presented here, we therefore expect to be able to use those abstractions, the activation SOMs, as a way to explore and manipulate conceptual knowledge as represented in neural network. We are in particular looking at how the comparison of multiple SOMs for multiple concepts can help identify relations between those concepts. Another exciting way in which this work opens up new avenues for research is by using activation SOMs to identify the presence of unknown concepts in the network. Indeed, we have shown here that we can `query' the network for the presence and importance of the representation of existing concepts. However, a key question is whether, in training to make a particular prediction, the network could have identified particular properties of the considered objects that group them into meaningful but not yet identified concepts. Being able to recognize such a phenomenon could have obvious benefits in knowledge acquisition and extraction, especially in areas where understanding the mechanisms by which a prediction is made, the concepts that the network might have `discovered', is just as important as making the prediction itself (e.g., when applying machine learning to scientific research).

\balance

\begin{acks}
Initial ideas and methods that led to the ones presented in this paper were explored during the internship of Maxime Haurel in the author's research team (see \url{https://github.com/MHaurel/nn-concept-interpreter}).
\end{acks}

\bibliographystyle{ACM-Reference-Format}
\bibliography{main}



\end{document}